# Physics-Informed Neural Networks for Device and Circuit Modeling: A Case Study of NeuroSPICE

Chien-Ting Tung, *Member, IEEE*, and Chenming Hu, *Life Fellow, IEEE*

***Abstract*—We present NeuroSPICE, a physics-informed neural network (PINN) framework for device and circuit simulation. Unlike conventional SPICE, which relies on time-discretized numerical solvers, NeuroSPICE leverages PINNs to solve circuit differential-algebraic equations (DAEs) by minimizing the residual of the equations through backpropagation. It models device and circuit waveforms using analytical equations in time domain with exact temporal derivatives. While PINNs do not outperform SPICE in speed or accuracy during training, they offer unique advantages such as surrogate models for design optimization and inverse problems. NeuroSPICE's flexibility enables the simulation of emerging devices, including highly nonlinear systems such as ferroelectric memories.**

## I. INTRODUCTION

SPICE (Simulation Program with Integrated Circuit Emphasis) was developed at UC Berkeley as a general-purpose differential system solver [1]. Modern circuit simulators solve discretized differential and algebraic equations using well-established numerical methods [1], [2]. As device research expands, chips increasingly integrate emerging physical effects such as ferroelectric (FE) materials [3], photonic elements [4], [5], and three-dimensional thermal coupling in 3DICs [6] that require new device formulations and Multiphysics coupling not easily expressed or maintained in existing SPICE compact-model workflows.

On the other hand, artificial intelligence (AI) benefits from enhancing IC technology. AI or neural networks (NN) can learn the relationship between input and output data through training. However, NN can also be used without data using loss functions based on physics equations as a physics-informed neural network (PINN) [7]. PINN solves the differential equations by minimizing the residual in the loss function. The differential equations define the loss function, and no data is needed. Several studies have demonstrated PINN for applications such as power system simulation [8], [9] and transistor simulation [10]. So far, there has been no study applying PINN to circuit simulations.

This work was supported by the Berkeley Device Modeling Center, University of California at Berkeley, Berkeley, CA, USA. The review of this article was arranged by Editor XXX. (Corresponding author: Chien-Ting Tung.)

The authors are with the Department of Electrical Engineering and Computer Sciences, University of California at Berkeley, Berkeley, CA 94720 USA (e-mail: cttung@berkeley.edu).

In this paper, we investigate the feasibility of PINN as circuit simulator. NeuroSPICE is a physics-informed neural network (PINN) framework that represents circuit unknowns (node voltages and selected branch currents) as continuous functions of time by taking time as an explicit network input. During training, the network outputs are differentiated with respect to time via backpropagation to obtain analytical time derivatives. This continuous, differentiable representation avoids finite-difference time approximations during loss evaluation and makes it straightforward to express device and Multiphysics equations inside a single Python environment. NeuroSPICE lowers the barrier to rapid prototyping of emerging device models (no Verilog-A required), naturally accommodates Multiphysics or higher-order formulations. It can be potentially used as differentiable circuit surrogate model for design optimization.

## II. METHODOLOGY

NeuroSPICE is formulated as a physics-informed neural network (PINN) (Fig. 1). The network input is time $t$ and the network outputs are node voltages (and, where desired, branch currents); the input may be extended to include design parameters or spatial coordinates for Multiphysics problems beyond conventional SPICE simulation. NeuroSPICE replaces numerical time stepping with a neural network that directly represents the entire waveform as an analytical function of time. During training, a set of time points spanning the simulation interval is presented to the network as the input vector, and the network returns an array of voltages (or currents) evaluated at those time points. These output waveforms are therefore represented by the neural network as continuous, analytical functions of time. The loss function is constructed from the residuals of the circuit differential-algebraic equations (DAEs) together with initial-condition constraints; an optimizer updates the network parameters to minimize this loss driving the DAE residuals and initial-condition errors toward zero.

Fig. 2 illustrates a representative DAE for a transistor amplifier, formed by applying Kirchhoff's current law at each node. The MOSFET model is implemented in Python as shown in Fig. 3. Unlike typical SPICE device models (for example, BSIM models written in Verilog-A [11], [12]), which are evaluated inside a simulator with limited user control, NeuroSPICE implements device models in Python to facilitate rapid prototyping. Device models may return currents, charges, or other quantities beyond the conventional voltage–current interface. A key feature of NeuroSPICE is that

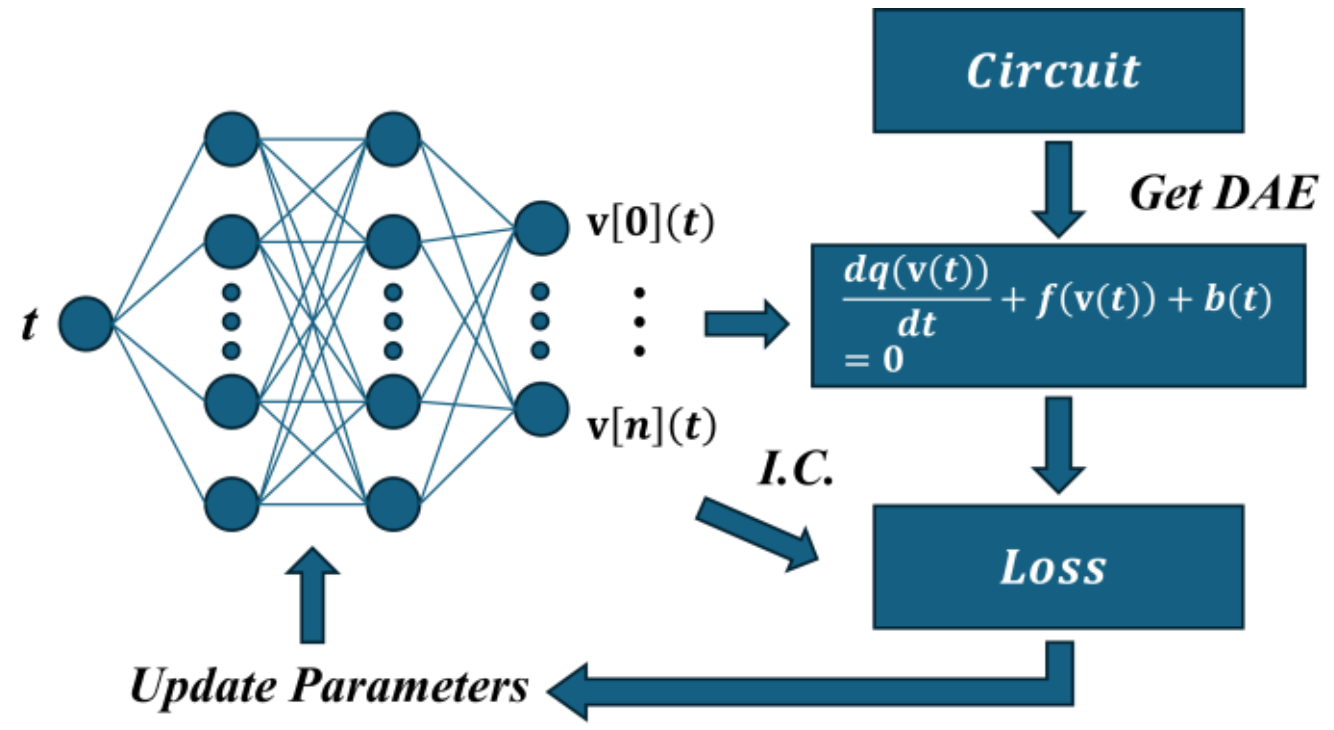

**Fig. 1.** The schematic diagram of NeuroSPICE. The loss function is the DAE of the circuit and the initial condition. The neural network will update its parameters to minimize loss. I.C. is the initial condition.

**DAE of a Transistor Amplifier**

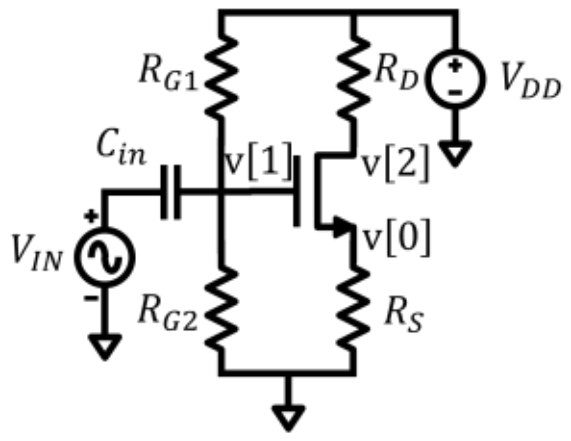

$$L0 = \frac{v[0]}{R_S} + nmos(v[2], v[1], v[0])[2]$$

$$L1 = \frac{v[2] - V_{DD}}{R_D} + nmos(v[2], v[1], v[0])[0]$$

$$L2 = C_{in}\frac{d(v[1] - V_{IN})}{dt} + \frac{v[1]}{R_{G2}} + \frac{v[1] - V_{DD}}{R_{G1}} + nmos(v[2], v[1], v[0])[1]$$

$$DAE_{loss} = Mean(L0^2 + L1^2 + L2^2) \qquad \frac{d}{dt} \text{ is computed by autograd}$$

$$Loss = \alpha \times IC_{loss} + \beta \times DAE_{loss}$$

**Fig. 2.** Example DAE of a transistor amplifier and its loss function. In this case, the KCL of each node will be solved including the current of the nmos. The time derivative is computed by the autograd function during backpropagation.

**Device Model Pseudocode**

```
class MOSFET:
    Parameter Initialization: ...
    def current(vd, vg, vs):
        vgs = vg − vs
        vgd = vg − vd
        Id = f1(vgs, vds)
        Ig = f2(vgs, vds)
        Qg = g1(vgs, vds)
        Qd = g2(vgs, vds)
        dQg/dt = autograd(Qg)        Use autograd to compute analytical derivative
        dQd/dt = autograd(Qd)
        return [Id + dQd/dt, Ig + dQg/dt, −Id − Ig − dQd/dt − dQg/dt]
```

**Fig. 3.** The pseudocode of the device model. The MOSFET DC and switching currents are computed and returned to the loss function. Charge currents (dQ/dt) are analytically computed by autograd.

these temporal derivatives such as dV/dt and dQ/dt (Fig. 3) are obtained analytically via automatic differentiation (autograd) provided by machine-learning frameworks such as PyTorch [13]. Because the neural network represents nodal voltages as explicit functions of the input time and autograd applies the chain rule during back propagation, dV/dt, dQ/dt, and related quantities are computed exactly from the network's analytical representation rather than by finite-difference approximations. Consequently, NeuroSPICE does not require time-domain discretization, linearization, or numerical integration schemes (e.g., backward Euler) used in conventional transient SPICE solvers. All required derivatives are available analytically during both training and inference. This allows easy implementation of emerging device models and Multiphysics models without knowing the numerical details in conventional SPICE. The total loss is a weighted sum of the DAE residual loss ($DAE_{loss}$) and the initial-condition loss ($IC_{loss}$), with coefficients a and b chosen to balance these terms [14].

## III. Result & Discussion

NeuroSPICE is implemented with Pytorch, Adam optimizer, and Tanh activation function. All cases use a 4-layer fully connected neural network with 50 neurons in each hidden layer. Training is done on a NVIDIA Quadro Pro GPU.

We evaluate NeuroSPICE with some simple MOSFET circuits. The transient response of the transistor amplifier in Fig. 2 is shown in Fig. 4, and HSPICE is used as a reference. The NeuroSPICE-calculated $V_D$ and $V_G$ closely match the HSPICE results. Fig. 5 shows a more complex five-stage ring oscillator; these results demonstrate that NeuroSPICE can simulate unstable, self-oscillating systems and solve DAEs that include multiple nonlinear device models.

NeuroSPICE is also a flexible platform for emerging devices and systems. As an example, FE devices are a promising technology for low-power in-memory computing. However, it is difficult to implement a well-posed ferroelectric compact model in SPICE simulators since it requires not only deep understanding of ferroelectric physics but also the knowledge of Verilog-A coding and circuit simulations [3]. The Landau–Khalatnikov (LK) model [15] is commonly used to describe FE material dynamics which is highly nonlinear. In NeuroSPICE, it is straightforward to implement this model in DAE form. Fig. 6 demonstrates NeuroSPICE's flexibility by simulating a FeRAM cell that couples the LK FE model with the MOSFET model. NeuroSPICE reproduces the voltage drop associated with polarization switching in the FE capacitor.

Table I summarizes the hyperparameters used for each case. We observe that highly nonlinear devices and circuits require more training epochs. In the FeRAM example, the LK model necessitated a larger number of epochs and a smaller learning rate to reach convergence. The training time for NeuroSPICE is longer than that of commercial SPICE, primarily because Python implementations are slower than optimized C/C++ code and because the PINN's internal matrices are typically larger than the circuit's nodal matrices. Thus, we do not think NeuroSPICE is a direct replacement for SPICE in normal circuit simulations. Despite the longer training time, NeuroSPICE inference is comparable to or faster than SPICE (approximately 200 µs in our tests). Because NeuroSPICE represents node voltages and branch quantities as continuous, differentiable functions of time and computes exact temporal derivatives via automatic differentiation, the framework is fully differentiable; by training NeuroSPICE with design parameters, NeuroSPICE can serve as a differentiable circuit surrogate in inference for inverse design and gradient-based optimization, with analytically represented output waveforms.

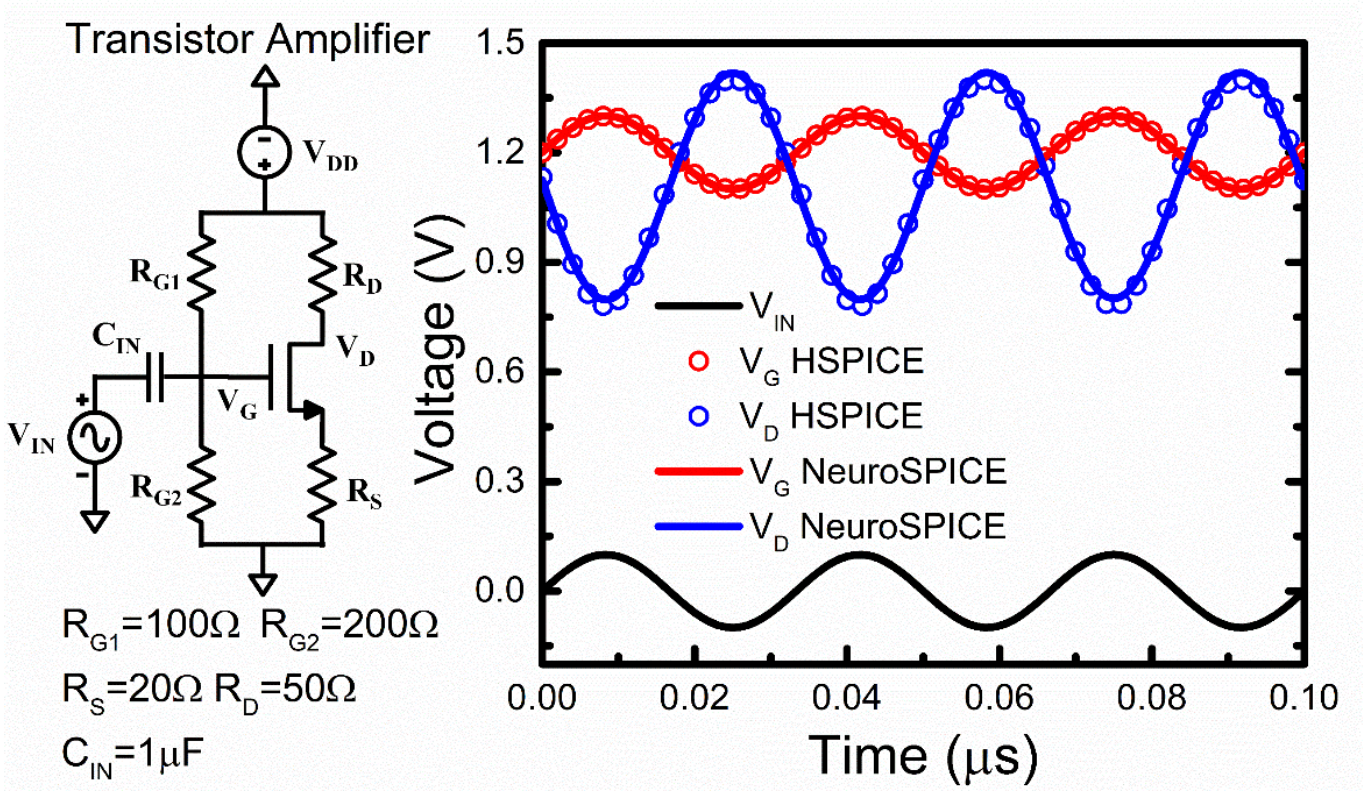

**Fig. 4.** The simulation of a transistor amplifier.

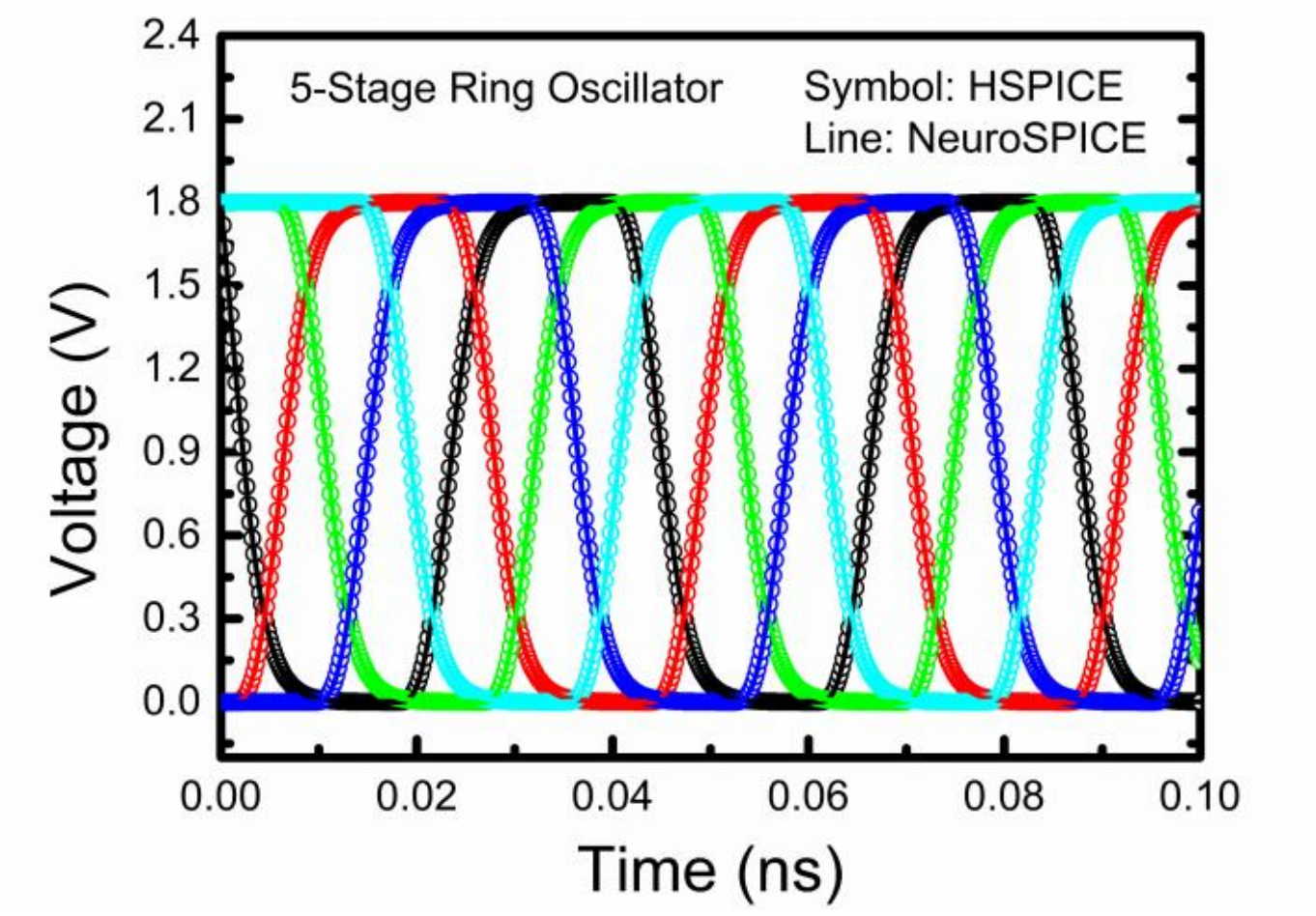

**Fig. 5.** The simulation of a 5-stage ring oscillator.

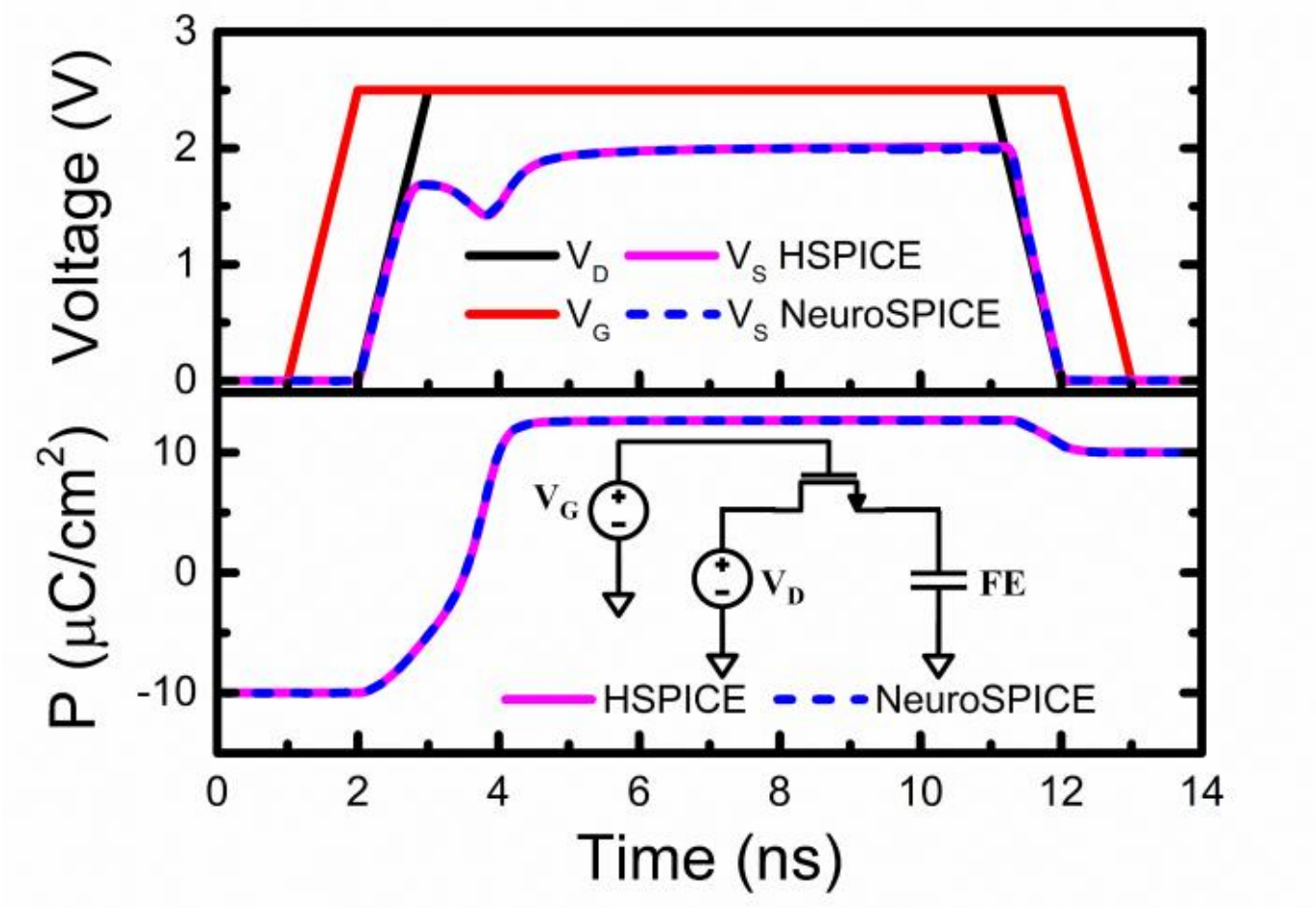

**Fig. 6.** The simulation of a FeRAM using the LK model.

TABLE I

Epochs, learning rate (LR), and training time for each case

| | Epochs | LR | Training time |
|---|---|---|---|
| Transistor amplifier | 25000 | 5e-3 | 4 min |
| Ring Oscillator | 20000 | 5e-3 | 7.21 min |
| FeRAM | 60000 | 2e-4 | 6.65 min |

The inference time is about 200 μs in all cases.

## IV. Conclusion

We study NeuroSPICE, a PINN-based circuit simulator that transcends the numerical time-stepping of conventional SPICE. It can solve circuits with strongly nonlinear devices and extend to Multiphysics systems for emerging-technology exploration. We have identified current challenges for NeuroSPICE and propose its potential usage as circuit surrogate models. Further work should evaluate scalability, convergence and its use in gradient-based design optimization.